\documentclass[11pt,a4paper]{article}
\usepackage[hyperref]{naaclhlt2018}
\usepackage{times}
\usepackage{latexsym}
\usepackage{amsmath}
\usepackage{amssymb}
\usepackage{pifont}
\usepackage{bm}
\usepackage{graphicx}
\usepackage{tikz}
\usepackage{ccg}
\usepackage{arydshln}
\usepackage{multirow}
\usepackage{bbm}
\usepackage{algorithm, algpseudocode}
\usetikzlibrary{positioning,decorations.pathmorphing,decorations.pathreplacing,shapes}
\newcommand{\argmax}{\mathop{\rm arg~max}\limits}
\newcommand{\cmark}{\ding{51}}
\newcommand{\xmark}{\ding{55}}

\usepackage{url}

\algtext*{EndFunction}
\algtext*{EndIf}
\algtext*{EndFor}
\algnewcommand{\LineComment}[1]{\State \(\triangleright\) #1}
\makeatletter
\renewcommand{\ALG@beginalgorithmic}{\small}
\makeatother
\aclfinalcopy % Uncomment this line for the final submission
\def\aclpaperid{833} %  Enter the acl Paper ID here

\title{Consistent CCG Parsing over Multiple Sentences \\ for Improved Logical Reasoning}

\author{Masashi Yoshikawa$^1$\\ {\tt yoshikawa.masashi.yh8@is.naist.jp}\\
        \And
        \hspace{1.5cm}Koji Mineshima$^2$\\\hspace{1.5cm}{\tt mineshima.koji@ocha.ac.jp}\\
        \AND
        Hiroshi Noji$^3$ \\ {\tt hiroshi.noji@aist.go.jp}\\
        \And
        \hspace{1.5cm}Daisuke Bekki$^2$\\\hspace{1.5cm}{\tt bekki@is.ocha.ac.jp}\\
        \AND
       $^1${\rm Nara Institute of Science and Technology, Nara, Japan} \\
       $^2${\rm Ochanomizu University, Tokyo, Japan} \\
       $^3${\rm Artificial Intelligence Research Center, AIST, Tokyo, Japan}
}

\date{}

\begin{document}
\maketitle
\begin{abstract}
    In formal logic-based approaches to Recognizing Textual Entailment (RTE),
    a Combinatory Categorial Grammar (CCG) parser is used to parse input premises and hypotheses to obtain their logical formulas.
    Here, it is important that the parser processes the sentences consistently;
    failing to recognize a similar syntactic structure
    results in inconsistent predicate argument structures among them,
    in which case the succeeding theorem proving is doomed to failure.
    In this work, we present a simple method to extend an existing CCG parser
    to parse a set of sentences consistently, which is achieved with
    an inter-sentence modeling with Markov Random Fields (MRF).
    % The decoding method based on dual decomposition guarantees the optimality
    % while not harming the efficiency of the original parser.
    When combined with existing logic-based systems,
    our method always shows improvement in the RTE experiments
    on English and Japanese languages.
\end{abstract}

\section{Introduction}
% 文のモデルは多いが文関を超えたモデルはまだ少ない,
% Neural network consitent 
% coferenceとかはあつかってるが
% syntacticなこともmulti sententialな問題のなんか
% PPアタッチメント(望遠鏡) ??
% RTEは同じ状況の言い換え
While today's neural network-based syntactic parsers~\cite{dyer-rnng:16,DBLP:journals/corr/DozatM16,yoshikawa-noji-matsumoto:2017:Long}
% part-of-speech taggers and syntactic parsers
have proven successful on sentence level modeling,
it is still challenging to accurately process texts that go beyond a single sentence
% there are still performance gap in processing texts that go beyond a single sentence
(e.g. coreference resolution, discourse structure analysis).
In this work we focus, among others, on the consistent analysis of multiple sentences in a document.
This is as an important problem in reasoning tasks as other document analysis.

% RTEはmultiple sentence の意味的な解析にとって要素的な技術である
RTE is an elemental technology for semantic analysis of multiple sentences,
% In particular, in RTE tasks, given a text (T) and a hypothesis (H), a system determines if T entails H.
where, given a text (T) and a hypothesis (H), a system determines if T entails H.
Existing methods based on formal logic~\cite{Bos:2008:WSA:1626481.1626503,martinezgomez-EtAl:2017:EACLlong,abzianidze:2017:EMNLP2017Demos}
obtain logical formulas for T and H using an off-the-shelf CCG parser, and then feed them to a theorem prover.
The standard approach to mapping CCG trees onto logical formulas is
to assign $\lambda$-terms to the words in a sentence and combine them in a bottom-up fashion~(Figure~\ref{noman}a).
% As their conversion to logical formulas follows
% the transparency of the surface syntax and the semantics of CCG,
% Naturally, when the parser fails to assign a correct syntactic structure, that error propagates to the entire system.
Here, when the parser fails to make consistent analyses for T and H,
the succeeding inference component is also doomed to failure.
In Figure~\ref{noman}b, when the parser wrongly analyzes ``{\it man exercising}''
in H as ``{\it man}'' modifying ``{\it exercising}'', % \footnote{This confuses even many syntactic analyzers today.},
% it is impossible to show T $\to$ H
the entailment relation cannot be established,
due to the different argument structures of {\tt exercise} in the resulting formulas.
% as it is impossible to show there is a man who has attribute ``exercise'' from T.
% as no event $e$ is in the resulting formula where a man is exercising.
% due to the wrongly assigned logical formulas.
% For these systems to work well on this kind of examples,
% a parser should be able to

\begin{figure}[t]
\centering
    \input noman.tex
\caption{
    (a) An example semantic template for verbs $V$ that
    associates a CCG category $S\backslash NP$ with a $\lambda$-term.
    % Terminal words are mapped
    % to $\lambda$-terms based on these templates.  % (a) The structure of a category is transparently translated to
    % the type of $\lambda$-term of its semantic template.
    % its use in a logical formula for ``There is no man exercising''.
    (b) A logical formula of a sentence is obtained at the root of a tree
    by composing $\lambda$-terms of all words following CCG combinatory rules.
    In this Figure, hypothesis H is wrongly parsed (See the text for details).
    }
% \caption{An example semantic template for a category $S\backslash NP$ and
%     its use in deriving a logical formula for ``There is no man exercising''.
%     A logical formula of a sentence can be obtained at the root of a tree
%     by composing $\lambda$-terms assigned to each word following CCG combinatory rules.
%     }
\label{noman}
\end{figure}

While it is ideal to enhance the overall performance of a parser, it is not cheaply obtainable.
Additionally, neural network-based parsers are susceptible to subtle changes in the input
and thus hard to inspect and modify its parameters to change its prediction.
Due to this, we cannot expect that a particular pair of words across multiple sentences
be always analyzed in a consistent manner.
% In NLP problems that involve more than one sentence,
% it is important to process the target examples in a consistent manner.

In this work, we solve the inconsistency problem above
by adapting the inter-sentence model of \citet{rush-EtAl:2012:EMNLP-CoNLL} to CCG parsing.
Their motivation is to exploit the similarities among test sentences
to overcome situations where the amount of the training data is scarce or its domain is different from the test data.
The method based on dual decomposition tries to find parse trees for a set of sentences
that agree with an MRF, which encourages the assignment of a similar structure to similar contexts.

% In this work, we solve the inconsistency problem discussed above
% by adapting Rush et al.'s method to the case of CCG parsing and logic-based approach to RTE.
% by building a global MRF that rewards the consistent CCG parsing across multiple sentences,
% Our motivation is different from theirs in that
In our approach, we aim to eliminate wrong logical formulas such as in Figure~\ref{noman}
by rewarding consistent CCG parses across sentences.
This, in turn, is achieved by rewarding the consistent assignment of categories
to the terminals. % , which is done by adapting the POS tagging of Rush et al.'s method.
This works for CCG parsing, as its derivation is mostly determined by the terminal categories.
% which we show is important in the experiments.
% Their method is simple and can be applied to a broad family of parsing algorithms
% while guarantees the optimality once a solution is found.
% They showed the effectiveness of their approach in dependency parsing and POS tagging.
% We solve the global decoding problem of CCG parser on sentences and MRF using dual decomposition.
% Specifically, we build a global MRF that rewards the consistent CCG parsing across multiple sentences.
The key of our approach is that by combining A* parsing of \citet{yoshikawa-noji-matsumoto:2017:Long}
with dual decomposition, we can keep small the latency incurred by the use of the iterative algorithm.

We conducted experiments using two state-of-the-art logic-based systems
\cite{martinezgomez-EtAl:2017:EACLlong,abzianidze:2017:EMNLP2017Demos}
and two RTE datasets for English and Japanese languages.
Our method always shows improvement compared to the baselines.

\section{Method}
% We describe our approach of
% joint modeling of CCG trees $Y=\langle {\bm y}_1, \ldots, {\bm y}_N \rangle$\footnote{
%     We denote a tuple of $a \in A$ and $b \in B$
%     as $\langle a, b \rangle \in A \times B$.
%     For $a_1, \ldots, a_N \in A$, we denote
%     ${\bm a} = \langle a_1, \ldots, a_N \rangle \in A \times \cdots \times A = A^N$.
% }
% for sentences $X=\langle {\bm x}_1, \ldots, {\bm x}_N \rangle$
% with the inter-consistency model expressed with MRFs.
We describe our approach of
modeling the inter-consistencies among CCG trees
$Y=\langle {\bm y}_1, \ldots, {\bm y}_N \rangle$
% \footnote{
%     We denote a tuple of $a \in A$ and $b \in B$
%     as $\langle a, b \rangle \in A \times B$.
%     For $a_1, \ldots, a_N \in A$, we denote
%     ${\bm a} = \langle a_1, \ldots, a_N \rangle \in A \times \cdots \times A = A^N$.
% }
for sentences $X=\langle {\bm x}_1, \ldots, {\bm x}_N \rangle$ (\S\ref{doc}),
\footnote{
    In this work, we focus on the inconsistency problem of premises and hypotheses of RTE task, and thus
    $X$ does not contain sentences from any ``training data'', as was done in \citet{rush-EtAl:2012:EMNLP-CoNLL}.
    Exploiting external resources in the same manner is also an interesting future direction.
}
A* parsing method for each ${\bm y}_i$ (\S\ref{astar}) and
joint decoding of the MRF and A* parsing using dual decomposition (\S\ref{joint}).

\subsection{Document Consistencies with MRF}
\label{doc}
\begin{figure}[t]
\centering
    % \documentclass[dvipdfmx,11pt]{standalone}
% \usepackage{tikz}
% \usetikzlibrary{positioning,decorations.pathmorphing,decorations.pathreplacing,shapes}
% %
% \begin{document}
%
  \begin{tikzpicture}
    \tikzset{block/.style={draw,rectangle,line width=0.8pt,fill=white!10,text centered,rounded corners,minimum height=0.6cm}};
      \node[] (h) {{\bf H}:};
      \node[right=0.1cm of h]  (h1) {There};
      \node[block, right=0.5cm of h1] (h2) {is};
      \node[right=0.5cm of h2] (h3) {no};
      \node[block, right=0.5cm of h3] (h4) {man};
      \node[block, right=0.5cm of h4] (h5) {exercising};

      \node[below=0.4cm of h] (t) {{\bf T}:};
      \node[right=0.5cm of t]  (t1) {A};
      \node[block, right=0.7cm of t1] (t2) {man};
      \node[block, right=0.7cm of t2] (t3) {is};
      \node[block, right=0.7cm of t3] (t4) {exercising};

      \node[draw, circle, line width=0.8pt, above left=0.8cm and -0.15cm of h5] (c1) {c3};
      \node[draw, circle, line width=0.8pt, left=0.7cm of c1] (c2) {c2};
      \node[draw, circle, line width=0.8pt, left=0.6cm of c2] (c3) {c1};
      \draw[-, line width=0.55pt, shorten <= 0.2pt, shorten >=0.5pt] (t4) -- (c1);
      \draw[-, line width=0.55pt, shorten <= 0.2pt, shorten >=0.5pt] (h5) -- (c1);
      \draw[-, line width=0.55pt, shorten <= 0.2pt, shorten >=0.5pt] (h4) -- (c2);
      \draw[-, line width=0.55pt, shorten <= 0.2pt, shorten >=0.5pt] (t2) -- (c2);
      \draw[-, line width=0.55pt, shorten <= 0.2pt, shorten >=0.5pt] (h2) -- (c3);
      \draw[-, line width=0.55pt, shorten <= 0.2pt, shorten >=0.5pt] (t3) -- (c3);
      % \draw[densely dashed, line width=0.55pt, shorten <= 0.8pt, shorten >=0.8pt] (h.north east) -- (t.south east);
  \end{tikzpicture}

% \end{document}
\caption{
    % An MRF for an example RTE problem,
    % where rectangles represent word nodes and circles
    % context nodes. We model the consistencies across sentences
    % through the interdependencies expressed with the edges.
    An MRF graph is made up of cliques each consisting of one {\it context node} ($\in C$; circles)
    and {\it word nodes} ($\in W$; rectangles) instantiating that context.
    As such, each clique expresses the interdependencies among words appearing across sentences.}
\label{mrf}
\end{figure}
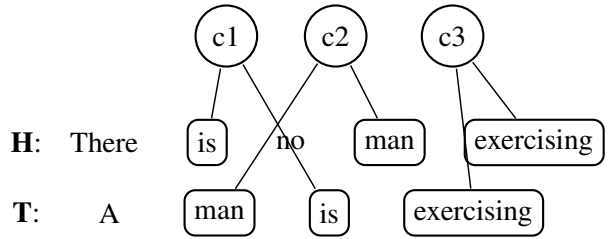

To model inter-consistencies among CCG parses,
we adapt the global MRF model of \citet{rush-EtAl:2012:EMNLP-CoNLL}.
See Figure~\ref{mrf} for an example MRF.
Our MRF encourages the assignment of similar categories
to the words appearing in similar contexts. % (connected words in the figure).
% This, in turn, encourages globally consistent CCG parsing. % , as the tree structure is mostly determined by the terminal categories.
% The MRF enforces soft constraints over sentences
% so that the words appearing in similar contexts are assigned similar categories.
% This in turn the logical forms obtained from the trees have the consistent predicate argument structure.
% Though this simple method only considers the terminals of parse trees,
% this works well for CCG parsing, as the tree structure is mostly determined by the terminal categories.

Firstly we construct a graphical representation of an MRF.
For each context~(unigram surface form in the case of Figure~\ref{mrf}) $c \in C$,
we have a set $W_c$ of indices $\langle s, t \rangle$ that appear in $c$,
where $s$ is a sentence index and $t$ a word index on sentence $s$.
% The elements of the set share the same contextual characteristics represented by $c$.
Let $W = \bigcup_{c \in C} W_c$. We define an undirected graph $G = \langle V, E \rangle$,
whose vertices are $V = C \cup W$ and edges $E = \{\langle w, c \rangle : c \in C, w \in W_c\}$.
% whose elements share the same contextual characteristics represented by $c$.
% where $s$ is a sentence index and $i$ is a word position in $s$.
See Figure~\ref{mrf} for an MRF graph constructed for an example RTE problem.
% In the following, nodes in $C$ are referred to as {\it context nodes},
% while nodes in $S = \bigcup_{c \in C} S_c$ as {\it word nodes}.
% Using these sets, we construct an undirected graph $G = \langle V, E \rangle$,
% whose vertices are $V = C \cup S$ and edges $E = \{(i, c) : c \in C, i \in S_c\}$.

% We tackle a problem of assigning to each node a label from the set of CCG categories ${\mathcal T}$,
% so as to maximize the score $g$, while considering the expressed interdependencies.
We assign to each node in the graph a label from a set of CCG categories ${\mathcal T}$,
so as to maximize the global consistency score $g$.
By combining $g$ with local CCG parsing for each ${\bm y}$,
we aim to obtain globally consistent trees $Y$ (\S\ref{joint}).
We define label assignment ${\bm z}$ to nodes in $V$ as
${\bm z} = \langle z_1, ..., z_{|W|}, z'_1,
..., z'_{|C|} \rangle \in {\mathcal T}^{|W|} \times {\mathcal T'}^{|C|}$,
% ..., z_C \rangle \in {\mathcal T} \times \cdots \times {\mathcal T}
% \times {\mathcal T'} \times \cdots \times {\mathcal T'}$,
where ${\mathcal T'} = {\mathcal T} \cup \{NULL\}$.
In the following, $z_w$ denotes the element in ${\bm z}$ at the index corresponding to $w \in W$
(similarly $z'_c$ for $c \in C$).
Following \citet{rush-EtAl:2012:EMNLP-CoNLL},
we allow $NULL$ label for context nodes. This works as a switch to ``turn off''
the consistency constraints to the connected nodes.
Then, in the set ${\mathcal Z}(X)$ of all possible ${\bm z}$s for $X$,
we look for ${\bm z}^* = \argmax_{{\bm z} \in {\mathcal Z}(X)} g({\bm z})$, where $g({\bm z})$ is\footnote{
    We omit unary terms $f_c$ for $c \in C$, as we set them 0.}:
% , on
% the nodes in the undirected graph $\langle V, E \rangle$,
% where $V = \{1, \ldots, C\} \cup (\bigcup_{c=1}^C S_c)$ and
% $E = \{(i, c) : c \in C, i \in S_c\}$.
% For context nodes, we assign a label from the set
% $L_c = {\mathcal T} \cup \{NULL\}$,
% where ${\mathcal T}$ stands for the set of all CCG supertags.
% For other nodes for words, we choose a supertag from $L_{(s, i)} = {\mathcal T}$.
% The score $g$ is a simple linear function:
\begin{align}
    g({\bm z}) = \sum_{w \in W} f_w(z_w) + \sum_{(w, c) \in E} f_{w,c}(z_w, z'_c). \nonumber
\end{align}
To reward the consistent assignment of categories among connected nodes,
$f_{w,c}$ is defined as follow:
\begin{align}
    & f_{w,c}(z_w, z'_c) = \begin{cases}
      \delta_1 & \text{if}\; z_w = z'_c \\
      \delta_2 & \text{if}\; {\tt simpl}(z_w) = {\tt simpl}(z'_c) \\
      \delta_3 & \text{if}\; z'_c = NULL \nonumber \\
      0        & \text{otherwise}, \nonumber \\
  \end{cases}
\end{align}
where $\delta_1 \geq \delta_2 \geq \delta_3$
and ${\tt simpl}$ removes feature values from a category
(e.g. ${\tt simpl}(S_{dcl}\backslash NP) = S\backslash NP$).
for $f_w$, we use $\log P_{tag}$ obtained by CCG parser (\S\ref{astar}).
We tune $\delta_i$s based on the RTE performance on the development set.

Since the above MRF $g({\bm z})$ has a simple na{\"i}ve Bayes
structure, we can compute $argmax$ using dynamic programming.

% This encourages that the connected context and
% word nodes are assigned the same label,
% which, in turn enforces that the words in the similar contexts
% have the same supertag.

\subsection{A* CCG Parsing}
\label{astar}
To parse a sentence, we use the state-of-the-art A* parsing method of~\citet{yoshikawa-noji-matsumoto:2017:Long},
which treats a CCG tree ${\bm y}$ as a tuple $\langle {\bm c}, {\bm h} \rangle$ of
categories ${\bm c}=\langle c_1, \ldots, c_M \rangle$
and dependency structure ${\bm h}=\langle h_1, \ldots, h_M \rangle$,
where each $h_i$ is a head index.
They model a tree with a locally factored model;
the probability of a CCG tree is the product of
the probabilities of the categories $p_{tag}$ and the dependency heads $p_{dep}$ of all words in ${\bm x}$:
\begin{align}
    \label{astarmodel}
    p({\bm y} | {\bm x}) & = \prod_{i \in [1,M]} p_{tag}(c_i | {\bm x}) \prod_{i \in [1,M]} p_{dep}(h_i | {\bm x}) \nonumber.
\end{align}
% For the detailed algorithm please refer to \citet{yoshikawa-noji-matsumoto:2017:Long}.
Note that the most computationally heavy part of their method is the calculation of ${P_{tag|dep}}$,
which needs to be done only once in our extension with dual decomposition.
The additional computational cost of our method is rather small, as
it depends on the number of times to run A* algorithm on the precomputed ${P_{tag|dep}}$,
which is quite efficient.\footnote{
    The supertagger of depccg processes 54 sentences per second
    while its A* decoder 2463 sentences per second.
    This is measured on SICK test set consisting of 9854 sentences
    using 2.20 GHz Intel Xeon CPUs with 16 cores.
}
% while A* decoding is highly efficient.
% This is convenient to us, as we extend their method with dual decomposition
% Note that once ${P_{tag|dep}}$ probabilities are computed,
% A* decoding can be efficiently run multiple times on the same input text.
% This is helpful when running the decoder in iterations of dual decomposition.

The probability $P(Y|X)$ of parses $Y$ for $X$ under this model is
simply the product of all ${\bm y}_i$s:
% The decoding problem
% can be solved by decomposing it into subproblems:
\begin{align}
    Y^* & = \argmax_{Y \in {\mathcal Y}(X)} P(Y | X) \nonumber \\
    & = \argmax_{Y \in {\mathcal Y}(X)} \sum_{{\bm y}_i \in Y} \log p({\bm y}_i | {\bm x}_i), \nonumber
\end{align}
where ${\mathcal Y}(X)$ is the space of all possible parses for $X$.

\subsection{Dual Decomposition}
\label{joint}
% Following \citet{rush-EtAl:2012:EMNLP-CoNLL},
To obtain CCG parses $Y$ for sentences $X$ that are optimal
in terms of both the global consistency model~(\S\ref{doc}) and the local parsing model~(\S\ref{astar}),
we solve the following problem using dual decomposition:
% We obtain CCG trees $Y$ for a document $X$ by solving the following problem with dual decomposition:
\begin{align}
    (Y^*, {\bm z}^*) & = \argmax_{Y \in {\mathcal Y}(X), {\bm z} \in {\mathcal Z}(X)}
    P(Y | X) + g({\bm z}) \nonumber \\
    & \text{s.t.} \; \forall  \langle s, t \rangle \in W\; z_{s, t} = c_{s, t}, \nonumber % \label{cond}
\end{align}
where $c_{s, t}$ is the category assigned on $t$'th word in ${\bm y}_s$.
The condition in the equation states that the decoded $Y^*$ and ${\bm z}^*$ must
agree in the category assignment to word nodes in the MRF.
Alg.~\ref{alg:dual} shows the pseudocode for dual decomposition applied to our method.
Note that all the decoding subproblems can be
kept intact even when added the Lagrangian multiplier $u$ of dual decomposition.
% Due to the space limitation, the algorithm of dual decomposition is omitted in this paper.
% Please refer to \citet{rush-EtAl:2012:EMNLP-CoNLL}.

\begin{algorithm}[t]
  \begin{algorithmic}
      % \State $\langle V, E \rangle = \langle S\cup C, E \rangle$: undirected graph for MRF (\S\ref{doc}),
      \LineComment $J$: a set of pairs of word nodes and categories in MRF
      \LineComment $\alpha$: step size ($0.0 < \alpha \leq 1.0$)
      \State Let $J = \{\langle w, c \rangle | w \in W, c \in {\mathcal T}\}$
      \State Let $\mathbbm{1}_c(z) = 1 \; \text{if} \; z \; \text{equals to} \; c \; \text{else} \; 0$
      \State $u_{w,c}^{(1)} \gets 0 \;\; \forall \langle w,c \rangle \in J$ 
      \For{$k = 1, \ldots, K$}
        \State ${\bm z}^{(k)} \gets  \argmax_{{\bm z} \in {\mathcal Z}(X)} g({\bm z}) + {\displaystyle \sum_{\langle w, c \rangle \in J}} u_{w,c}^{(k)} \mathbbm{1}_c(z_w)$
        % \Comment \S\ref{doc}
        \State $Y^{(k)} \gets \argmax_{Y \in {\mathcal Y}(X)} P(Y | X) - {\displaystyle \sum_{\langle w,c \rangle \in J}} u_{w,c}^{(k)} \mathbbm{1}_c(c_w)$
        % \Comment \S\ref{astar}
        \If {$z_w^{(k)} = c_w^{(k)} \; \text{for all} \; w \in W$}
          \State \Return $\langle {\bm z}^{(k)}, Y^{(k)} \rangle$
        \EndIf
        % \For{$\langle c, i \rangle \in J$}
          \State $u_{w,c}^{(k+1)} \gets u_{w,c}^{(k)} + \alpha (\mathbbm{1}_c(z_w^{(k)}) - \mathbbm{1}_c(c_w^{(k)}) ) \; \forall \langle w,c \rangle \in J$
        % \EndFor
      \EndFor
      \State \Return $\langle {\bm z}^{(K)}, Y^{(K)} \rangle$
  \end{algorithmic}
  \caption{Joint CCG parsing and global MRF decoding}
  \label{alg:dual}
\end{algorithm}

\section{Experiments}
\label{experiment}
% We conducted experiments to evaluate our method in terms of RTE performance.

\subsection{Experimental Settings}
\paragraph{English}
In English experiment, we test the performance of
ccg2lambda~\cite{martinezgomez-EtAl:2017:EACLlong} and LangPro~\cite{abzianidze:2017:EMNLP2017Demos}
on SICK dataset~\cite{MARELLI14.363.L14-1314}\footnote{
    We also conducted experiments on FraCaS dataset~\cite{fracas}.
    % For ccg2lambda, our parsing method (depccg with MRF) does not lead to any improved RTE results,
    % while for LangPro, two more problems are solved than when without MRF.
    For ccg2lambda, we found no improvements in RTE performance with our MRF, while for LangPro, we found that MRF guides to solve additional two problems.
}.
% As mentioned earlier, these systems use CCG parser to obtain logical formulas.
As mentioned earlier,
these systems try to prove whether T entails H, by applying a theorem prover
to the logical formulas converted from the CCG trees.
% apply a CCG parser to T and H texts,
% then convert the resulting trees to logical formulas by composing $\lambda$-terms for each word,
We report results for ccg2lambda with the default settings~(with SPSA abduction;~\citet{martinezgomez-EtAl:2017:EACLlong})
and results for two versions of LangPro, one which is described in \citet{abzianidze:2015:EMNLP} (henceforth we refer to it as LangPro15)
and the other in \citet{abzianidze:2017:EMNLP2017Demos} (LangPro17).\footnote{
    We report the scores for LangPro improved from the reviewed version,
    which we obtained from the author through the personal communication after the acceptance.
}
Briefly, the difference between the two versions is that LangPro17 is more robust to parse errors.
See the paper for the detail.
For the CCG parser in \S\ref{astar}, we use depccg\footnote{\label{depccg}\url{https://github.com/masashi-y/depccg}} with an MRF in \S\ref{doc}.
We compare our results with depccg without the MRF and
baselines reported in the above papers that use EasyCCG~\cite{lewis-steedman:2014:EMNLP2014}.

In MRF, a context node is constructed when two or more words from both T and H share the same surface form.
Exceptionally, some pairs of categories are allowed to be aligned with score $\delta_1$:
% Specifically, we choose the following category pairs:
a pair of noun modifier ($N/N$) and verb tense ($S_{ng}\backslash NP$),
which are categories for present participles, and a pair of
nominal modifier ($N/N$) and noun ($N$). In the experiment using ccg2lambda the pairs of categories of
transitive and intransitive verbs, ($(S_X\backslash NP)/NP$, $S_X\backslash NP$) and
($(S_X\backslash NP)/PP$, $S_X\backslash NP$), for any feature $X$ are also allowed with $\delta_1$.
% We construct an MRF node when two or more words from both T and H share the same surface form.

For the hyperparamters, we conducted grid search over $[0.0, 0.1, \ldots, 0.9]$ 
for each $\delta_i$ in the MRF s.t. $\delta_1 \geq \delta_2 \geq \delta_3$
and found that $\delta_1 = 0.9, \delta_2 = 0.1, \delta_3 = 0.0$
works the best on SICK trial set.
We set $\alpha = 0.0002$ and $K = 500$ in Alg.~\ref{alg:dual}.
We decay $\alpha$ by $0.9$ in every iteration.

\paragraph{Japanese}
In Japanese experiment, we evaluate ccg2lambda's performance
on JSeM dataset~\cite{Kawazoe2017}.
To construct an MRF graph, we processed RTE problems with kuromoji\footnote{\url{http://www.atilika.org/}}
and made a context node for a noun or a verb followed by an adverb.
% In contrast to English, more care must be taken here;
The reason why we use bigram POS tag-based context is that
the graph construction based on the surface form has
resulted in poor RTE performance, by overgenerating MRF constraints.
This may be due to the fact that Japanese sentences are usually tokenized into smaller units.
We used depccg and the same hyperparameters as English experiment.

\begin{table}[t]
    \centering
\scalebox{0.88}{
\begin{tabular}{lcccc} \hline
    \multicolumn{1}{c}{{\bf Method}} & {\bf Accuracy} & {\bf Precision} & {\bf Recall} \\ \hline
    \multicolumn{2}{l}{{\it LangPro15}~\cite{abzianidze:2015:EMNLP}}  \\
    ~~EasyCCG   &  79.05 & {\bf 98.00}  & 52.67 \\
    ~~depccg    &  80.37 & 97.94  & 55.81 \\
    ~~depccg + MRF  &  {\bf 80.88} & 97.91  & {\bf 57.03} \\ \hdashline
    \multicolumn{2}{l}{{\it LangPro17}~\cite{abzianidze:2017:EMNLP2017Demos}}  \\
    ~~EasyCCG   &  81.04 & 97.47  & 57.69 \\
    ~~depccg    &  81.53 & 97.51  & 58.81 \\
    ~~depccg + MRF  &  {\bf 81.61} & {\bf 97.52}  & {\bf 59.00} \\ \hdashline
    \multicolumn{3}{l}{{\it ccg2lambda}~\cite{martinezgomez-EtAl:2017:EACLlong}}   \\
    ~~EasyCCG   & 81.59 &  {\bf 97.73} &  58.48 \\
    ~~depccg   &  81.95 &      97.19  & 59.98 \\
    ~~depccg + MRF &  {\bf 82.86} & 97.14  & {\bf 62.18} \\ \hline
\end{tabular}}
    \caption{RTE results on test section of SICK}
\label{english_experiment}
\end{table}

% \begin{table}[t]
%     \centering
% \scalebox{0.73}{
% \begin{tabular}{cccccc} \hline
%     {\bf Method} & {\bf MRF} & {\bf Accuracy} & {\bf Precision} & {\bf Recall} \\ \hline
%     LangPro(EasyCCG) & \xmark  &  79.05 & {\bf 98.00}  & 52.67 \\ \hdashline
%     LangPro           & \xmark  &  78.85 & 97.48  & 52.48 \\
%     LangPro           & \cmark   &  {\bf 79.20} & 97.60  & {\bf 53.23} \\ \hline
%     ccg2lambda(EasyCCG) & \xmark  & 81.59 &  {\bf 97.73} &  58.48 \\ \hdashline
%     ccg2lambda & \xmark  &  81.95 &      97.19  & 59.98 \\
%     ccg2lambda & \cmark   &  {\bf 82.86} & 97.14  & {\bf 62.18} \\ \hline
% \end{tabular}}
%     \caption{RTE results on test section of SICK with inter-sentence
%     constraints on \cmark and off \xmark}
% \label{english_experiment}
% \end{table}

\begin{table}[t]
    \centering
\scalebox{0.88}{
\begin{tabular}{lccc} \hline
    \multicolumn{1}{c}{{\bf Method}} & {\bf Accuracy} & {\bf Precision} & {\bf Recall} \\ \hline
    ~~jigg &  75.0 & 92.7  & 65.4 \\
    ~~depccg &  67.87 & 88.34  & 56.77 \\
    ~~depccg + MRF &  71.31 & 88.88  & 62.24 \\ \hline
\end{tabular}}
    \caption{RTE results using ccg2lambda on JSeM}
\label{japanese_experiment}
\end{table}

% \begin{table}[t]
%     \centering
% \scalebox{0.78}{
% \begin{tabular}{ccccc} \hline
%     {\bf Method} & {\bf MRF} & {\bf Accuracy} & {\bf Precision} & {\bf Recall} \\ \hline
%     ccg2lambda(jigg) & \xmark  &  75.0 & 92.7  & 65.4 \\ \hdashline
%     ccg2lambda & \xmark  &  67.87 & 88.34  & 56.77 \\
%     ccg2lambda & \cmark  &  71.31 & 88.88  & 62.24 \\ \hline
% \end{tabular}}
%     \caption{RTE results on JSeM}
% \label{japanese_experiment}
% \end{table}

\begin{table*}[t]
    \centering
\scalebox{0.85}{
\begin{tabular}{cl} \hline
    & \multicolumn{1}{c}{{\bf Sentences}} \\ \hline
    \multirow{3}{*}{(a)}  &T: 
            The girl is sitting on the couch and is [$_{\small{S_{ng}\backslash NP}}$ crocheting] \\
       & H: The girl is sitting on the sofa and {\bf crocheting} \\
       & {\bf crocheting}: \xmark\; $N \; \leadsto$ \;\cmark\; $S_{ng} \backslash NP$ \\ \hdashline
    \multirow{4}{*}{(b)}  &T: 
           A veteran is showing different things {\bf from} a war {\bf to} some people \\
       & H: Different things [$_{\small{(\!NP\!\backslash \!NP\!)\!/\!NP}}$ from] a war are being shown
      [$_{\small{( \!( \!S \! \backslash \! NP \!) \!\backslash \! ( \!S \! \backslash \! NP \!) \!) \!/ \!NP}}$ to] some people by a veteran \\
       & {\bf from}: \xmark\; $((S \backslash NP)\backslash (S \backslash NP))/NP \;\leadsto$ \;\cmark\; $(NP\backslash NP)/NP$ \\
       & {\bf to}: \xmark\; $(NP\backslash NP)/NP \; \leadsto$ \;\cmark\; $((S \backslash NP)\backslash (S \backslash NP))/NP$ \\ \hdashline
    % \multirow{3}{*}{(a)} & \multirow{3}{*}{\small{3315}} &T: A woman is cooking a pork chop which is {\bf breaded} \\
    %    &     &H: A woman is cooking a breaded chop \\
    %   &  & {\bf breaded}: \xmark\; $S_{pss}\backslash NP \to$ \;\cmark\; $S_{adj}\backslash NP$ \\ \hdashline
    \multirow{3}{*}{(c)}  &T: A few man in a competition are [$_{\small{S_{ng}\backslash NP}}$ running] outside \\
       &H: A few man in a competition are {\bf running} outdoors \\
       & {\bf running}: \xmark \; $\small{(S_{ng}\backslash NP)/ NP} \; \leadsto$ \; \cmark
      $\small{S_{ng}\!\backslash\! NP}$ \\ \hdashline \hdashline
      % &  & {\bf outdoors}: \xmark \; $\small{N} \; \leadsto$ \; \cmark
      % $\small{(S\backslash NP)\backslash (S\backslash NP)}$ \\ \hdashline \hdashline
    \multirow{3}{*}{(d)}  &T: A man is [$_{\small{(S_{ng}\backslash NP)/NP}}$ eating] some food \\
       &H: The person is {\bf eating} \\
       & {\bf eating}: \cmark\; $S_{ng}\backslash NP \; \leadsto$ \;\xmark\; $(S_{ng}\backslash NP)/ NP$ \\ \hline

\end{tabular}
}
    \caption{Example parse results in SICK test set. (a), (b), (c) With the global MRF model,
    words in bold font previously assigned a wrong category (\xmark) have been assigned a correct one (\cmark).
    (d) is a case where the MRF is too strict and leads to the wrong assignment.}
\label{examples}
\end{table*}

\subsection{Results and Error Analysis}
We show the results on SICK in Table~\ref{english_experiment}.
% where the MRF column represents the use of an inter-sentence model.
Our MRF consistently contributes to the improvement of the accuracies for
both ccg2lambda and LangPro.
We observe the same tendency in the scores for all systems;
with MRF, both the accuracy and recall for the systems moderately improve and
the systems using depccg have higher recall and lower precision compared to the ones with EasyCCG
(with LangPro17 it marks higher precision as well).

In SICK, there are many instances of the construction shown in Figure~\ref{noman}
(``{\it There is no man exercising}'', ``{\it There is no dog barking}'', etc.),
whose correct reading is that the last verb (e.g. {\it exercising}) is a present participle modifying a noun (e.g. {\it man}).
EasyCCG and default depccg wrongly parse the last phrase ({\it man exercising}) as $N / N \; N$,
where {\it man} modifies {\it exercising}.
% Since another sentence provides less ambiguous analysis for ``{\it exercising}'',
Our method correctly predicts $N \; S_{ng} \backslash NP$, by utilizing
the paired sentence (e.g. ``{\it A man is exercising}''), in which the role of {\it exercising} is less ambiguous.
% our method utilizes that information and correctly predicts $N \; S_{ng} \backslash NP$.

% Table~\ref{examples} shows other interesting examples for which our method was successful.
Given that the strength of LangPro17 is its robustness to parse errors such as PP-attachment,
the larger gain in the accuracy for LangPro15 (roughly 0.5 versus 0.1 point up) indicates that
our method is also robust in handling well-known difficult parsing problems. % these constructions.
The example~(a) in Table~\ref{examples} is a case of coordinate construction.
Baseline depccg wrongly coordinates {\it crocheting} with a noun {\it sofa}, while
% With the presence of the same ``{\it crocheting}'' in T that works as a present particle,
our method successfully resolves the correct coordinate structure by
assigning $S_{ng} \backslash NP$ to the word (hence attaching it to {\it sitting}).
% changing the category for the former into the one for an intransitive verb.
Example~(b) is one of the cases of PP-attachment that our method successfully resolved.
Our method relocates the two PPs in T in their correct places.
As in the example in Figure~\ref{noman}, our method corrects cases like (a) and (b)
by using the structure of the less ambiguous counterpart as a guide.
% These examples show that some notoriously difficult parsing problems can be handled
% In example (a), all the other parsers wrongly assign a category for passive participle $S_{pss}\backslash NP$
% to ``{\it breaded}'', which actually is adjectival $S_{adj}\backslash NP$.
In the case of (c), the existing parsers misclassify {\it outdoors} in T as a noun
and turns the verb {\it run} into a transitive verb.
With our method, intransitive verb {\it run} in H works as a soft
constraint on the verb in T and corrects its structure successfully.
However, there are some cases where
using only surface forms as a cue forces the assignment of categories
which is consistent but not desirable.
% In example (d), {\it eat} in T and H are in transitive and intransitive uses, and thus should have different categories.
In example (d), {\it eat} is used as a transitive verb in T and as an intransitive verb in H; thus it should have different categories.

We show the results on JSeM in Table~\ref{japanese_experiment}.
The RTE performance for Japanese language has improved consistently across all the scores when we add an MRF.
However all the scores with depccg (with or without MRF) lag behind the scores
reported in~\citet{mineshima-EtAl:2016:EMNLP2016}, which uses a CCG parser implemented in Jigg~\cite{noji-miyao:2016:P16-4}.
We hypothesize that this is due to the fact that the previous work created the semantic templates for this language
by analyzing parse outputs by Jigg and this resulted in a kind of ``overfitting'' in the templates.

In the above experiments, our method worked well,
mainly due to the fact that the sentences in these datasets have comparably simple structure.
However, in other datasets, there are naturally more complex cases as in Table~\ref{examples}~(d), where
we want different syntactic analyses for occurences of words with the same surface form.
% we have intransitive use of ``{\it eat}''
% in one place and the transitive one in the other and do not want to force them into the same category.
% However, there are other datasets with more complex sentences where
% it is easily imagined that there are cases when we have intransitive ``{\it run}''
% in one place and the transitive one in the other and do not want to force them into the same category.
% In such a case, we can simply extend the definition of ``context'' by N-grams or the use of POS tag
We can counter these cases by simply extending the definition of ``context'' by N-grams or the use of POS tag
as we did in the Japanese experiment.
Developing a machine learning-based method that
selects which contexts to use and set $\delta_i$s automatically is also an important future work.
% It would be interesting to develop a machine learning-based system that
% selects which contexts to use and set $\delta_i$s automatically in future work.

\section{Conclusion and Future Work}
% In this work, we have shown that the problem of the inconsistent CCG parsing
% of multiple sentences can be solved with a document-level MRF and dual decomposition.
In this work, by modeling the inter-consistencies of multiple sentences in CCG parsing,
we have successfully improved the performance of the formal logic-based methods to RTE.
% In the experiments, our method consistently improves the performance of formal logic-based RTE systems.
Still, there can be pairs of words in more complex RTE problems
that should not have the same category but that our method wrongly force them to.
This is mainly due to the fact that we hand-tuned rules to construct context nodes.
In future work, we extend the method so that it learns when to set an MRF constraint.
%\bibliography{naaclhlt2018}

\section*{Acknowledgments}
First of all, we thank the three anonymous reviewers for their insightful comments.
We are also grateful to Lasha Abzianidze for conducting in-depth experiments and
for detailed discussion about LangPro.
This work was supported by JST CREST Grant Number JPMJCR1301, Japan.

\bibliography{naacl}

\begin{thebibliography}{}
\expandafter\ifx\csname natexlab\endcsname\relax\def\natexlab#1{#1}\fi

\bibitem[{Abzianidze(2015)}]{abzianidze:2015:EMNLP}
Lasha Abzianidze. 2015.
\newblock A tableau prover for natural logic and language.
\newblock In {\em Proceedings of the 2015 Conference on Empirical Methods in
  Natural Language Processing\/}. Association for Computational Linguistics,
  Lisbon, Portugal, pages 2492--2502.

\bibitem[{Abzianidze(2017)}]{abzianidze:2017:EMNLP2017Demos}
Lasha Abzianidze. 2017.
\newblock Lang{P}ro: {N}atural {L}anguage {T}heorem {P}rover.
\newblock In {\em Proceedings of the 2017 Conference on Empirical Methods in
  Natural Language Processing: System Demonstrations\/}. Association for
  Computational Linguistics, Copenhagen, Denmark, pages 115--120.

\bibitem[{Bos(2008)}]{Bos:2008:WSA:1626481.1626503}
Johan Bos. 2008.
\newblock {Wide-coverage Semantic Analysis with Boxer}.
\newblock In {\em Proceedings of the 2008 Conference on Semantics in Text
  Processing\/}. Association for Computational Linguistics, Stroudsburg, PA,
  USA, STEP '08, pages 277--286.

\bibitem[{Cooper et~al.(1996)Cooper, Crouch, Eijck, Fox, Genabith, Jaspars,
  Kamp, Milward, Pinkal, Poesio, Pulman, Briscoe, Maier, and Konrad}]{fracas}
Robin Cooper, Dick Crouch, Jan~Van Eijck, Chris Fox, Josef~Van Genabith, Jan
  Jaspars, Hans Kamp, David Milward, Manfred Pinkal, Massimo Poesio, Steve
  Pulman, Ted Briscoe, Holger Maier, and Karsten Konrad. 1996.
\newblock {FraCaS: A Framework for Computational Semantics}.
\newblock Deliverable D16.

\bibitem[{Dozat and Manning(2017)}]{DBLP:journals/corr/DozatM16}
Timothy Dozat and Christopher~D. Manning. 2017.
\newblock {Deep Biaffine Attention for Neural Dependency Parsing}.
\newblock {\em In Proc. of ICLR\/} .

\bibitem[{Dyer et~al.(2016)Dyer, Kuncoro, Ballesteros, and
  Smith}]{dyer-rnng:16}
Chris Dyer, Adhiguna Kuncoro, Miguel Ballesteros, and Noah~A. Smith. 2016.
\newblock {Recurrent Neural Network Grammars}.
\newblock In {\em Proceedings of the 2016 Conference of the North American
  Chapter of the Association for Computational Linguistics: Human Language
  Technologies\/}. Association for Computational Linguistics, San Diego,
  California, pages 199--209.

\bibitem[{Kawazoe et~al.(2017)Kawazoe, Tanaka, Mineshima, and
  Bekki}]{Kawazoe2017}
Ai~Kawazoe, Ribeka Tanaka, Koji Mineshima, and Daisuke Bekki. 2017.
\newblock An inference problem set for evaluating semantic theories and
  semantic processing systems for japanese.
\newblock In Mihoko Otake, Setsuya Kurahashi, Yuiko Ota, Ken Satoh, and Daisuke
  Bekki, editors, {\em New Frontiers in Artificial Intelligence: JSAI-isAI 2015
  Workshops, LENLS, JURISIN, AAA, HAT-MASH, TSDAA, ASD-HR, and SKL, Kanagawa,
  Japan, November 16-18, 2015, Revised Selected Papers\/}. Springer
  International Publishing, Cham, pages 58--65.

\bibitem[{Lewis and Steedman(2014)}]{lewis-steedman:2014:EMNLP2014}
Mike Lewis and Mark Steedman. 2014.
\newblock {A* CCG Parsing with a Supertag-factored Model}.
\newblock In {\em Proceedings of the 2014 Conference on Empirical Methods in
  Natural Language Processing (EMNLP)\/}. Association for Computational
  Linguistics, pages 990--1000.

\bibitem[{Marelli et~al.(2014)Marelli, Menini, Baroni, Bentivogli, bernardi,
  and Zamparelli}]{MARELLI14.363.L14-1314}
Marco Marelli, Stefano Menini, Marco Baroni, Luisa Bentivogli, Raffaella
  bernardi, and Roberto Zamparelli. 2014.
\newblock {A} {SICK} cure for the evaluation of compositional distributional
  semantic models.
\newblock In Nicoletta Calzolari, Khalid Choukri, Thierry Declerck, Hrafn
  Loftsson, Bente Maegaard, Joseph Mariani, Asuncion Moreno, Jan Odijk, and
  Stelios Piperidis, editors, {\em Proceedings of the Ninth International
  Conference on Language Resources and Evaluation (LREC'14)\/}. European
  Language Resources Association (ELRA), Reykjavik, Iceland, pages 216--223.
\newblock ACL Anthology Identifier: L14-1314.

\bibitem[{Mart\'{i}nez-G\'{o}mez et~al.(2017)Mart\'{i}nez-G\'{o}mez, Mineshima,
  Miyao, and Bekki}]{martinezgomez-EtAl:2017:EACLlong}
Pascual Mart\'{i}nez-G\'{o}mez, Koji Mineshima, Yusuke Miyao, and Daisuke
  Bekki. 2017.
\newblock {O}n-demand {I}njection of {L}exical {K}nowledge for {R}ecognising
  {T}extual {E}ntailment.
\newblock In {\em Proceedings of the 15th Conference of the European Chapter of
  the Association for Computational Linguistics: Volume 1, Long Papers\/}.
  Association for Computational Linguistics, Valencia, Spain, pages 710--720.

\bibitem[{Mineshima et~al.(2016)Mineshima, Tanaka, Mart\'{i}nez-G\'{o}mez,
  Miyao, and Bekki}]{mineshima-EtAl:2016:EMNLP2016}
Koji Mineshima, Ribeka Tanaka, Pascual Mart\'{i}nez-G\'{o}mez, Yusuke Miyao,
  and Daisuke Bekki. 2016.
\newblock {Building compositional semantics and higher-order inference system
  for a wide-coverage Japanese CCG parser}.
\newblock In {\em Proceedings of the 2016 Conference on Empirical Methods in
  Natural Language Processing\/}. Association for Computational Linguistics,
  Austin, Texas, pages 2236--2242.

\bibitem[{Noji and Miyao(2016)}]{noji-miyao:2016:P16-4}
Hiroshi Noji and Yusuke Miyao. 2016.
\newblock {Jigg: A Framework for an Easy Natural Language Processing Pipeline}.
\newblock In {\em Proceedings of ACL-2016 System Demonstrations\/}. Association
  for Computational Linguistics, pages 103--108.

\bibitem[{Rush et~al.(2012)Rush, Reichart, Collins, and
  Globerson}]{rush-EtAl:2012:EMNLP-CoNLL}
Alexander Rush, Roi Reichart, Michael Collins, and Amir Globerson. 2012.
\newblock {I}mproved {P}arsing and {POS} {T}agging {U}sing {I}nter-{S}entence
  {C}onsistency {C}onstraints.
\newblock In {\em Proceedings of the 2012 Joint Conference on Empirical Methods
  in Natural Language Processing and Computational Natural Language
  Learning\/}. Association for Computational Linguistics, Jeju Island, Korea,
  pages 1434--1444.

\bibitem[{Yoshikawa et~al.(2017)Yoshikawa, Noji, and
  Matsumoto}]{yoshikawa-noji-matsumoto:2017:Long}
Masashi Yoshikawa, Hiroshi Noji, and Yuji Matsumoto. 2017.
\newblock A* {CCG} {P}arsing with a {S}upertag and {D}ependency {F}actored
  {M}odel.
\newblock In {\em Proceedings of the 55th Annual Meeting of the Association for
  Computational Linguistics (Volume 1: Long Papers)\/}. Association for
  Computational Linguistics, Vancouver, Canada, pages 277--287.

\end{thebibliography}


\begin{thebibliography}{}
\expandafter\ifx\csname natexlab\endcsname\relax\def\natexlab#1{#1}\fi

\bibitem[{Aho and Ullman(1972)}]{Aho:72}
Alfred~V. Aho and Jeffrey~D. Ullman. 1972.
\newblock {\em The Theory of Parsing, Translation and Compiling\/}, volume~1.
\newblock Prentice-Hall, Englewood Cliffs, NJ.

\bibitem[{{American Psychological Association}(1983)}]{APA:83}
{American Psychological Association}. 1983.
\newblock {\em Publications Manual\/}.
\newblock American Psychological Association, Washington, DC.

\bibitem[{Chandra et~al.(1981)Chandra, Kozen, and Stockmeyer}]{Chandra:81}
Ashok~K. Chandra, Dexter~C. Kozen, and Larry~J. Stockmeyer. 1981.
\newblock \href{https://doi.org/10.1145/322234.322243}{Alternation}.
\newblock {\em Journal of the Association for Computing Machinery\/}
  28(1):114--133.
\newblock \url{https://doi.org/10.1145/322234.322243}.

\bibitem[{for Computing~Machinery(1983)}]{ACM:83}
Association for Computing~Machinery. 1983.
\newblock {\em Computing Reviews\/} 24(11):503--512.

\bibitem[{Goodman et~al.(2016)Goodman, Vlachos, and Naradowsky}]{P16-1001}
James Goodman, Andreas Vlachos, and Jason Naradowsky. 2016.
\newblock \href{https://doi.org/10.18653/v1/P16-1001}{Noise reduction and
  targeted exploration in imitation learning for abstract meaning
  representation parsing}.
\newblock In {\em Proceedings of the 54th Annual Meeting of the Association for
  Computational Linguistics (Volume 1: Long Papers)\/}. Association for
  Computational Linguistics, pages 1--11.
\newblock \url{https://doi.org/10.18653/v1/P16-1001}.

\bibitem[{Gusfield(1997)}]{Gusfield:97}
Dan Gusfield. 1997.
\newblock {\em Algorithms on Strings, Trees and Sequences\/}.
\newblock Cambridge University Press, Cambridge, UK.

\bibitem[{Harper(2014)}]{C14-1001}
Mary Harper. 2014.
\newblock \href{http://aclweb.org/anthology/C14-1001}{Learning from 26
  languages: Program management and science in the babel program}.
\newblock In {\em Proceedings of COLING 2014, the 25th International Conference
  on Computational Linguistics: Technical Papers\/}. Dublin City University and
  Association for Computational Linguistics, page~1.
\newblock \url{http://aclweb.org/anthology/C14-1001}.

\end{thebibliography}
\bibliographystyle{acl_natbib}
% ------------------------------------------------------
%  Gold\Prover       YES       NO       UNK       DEF
% ------------------------------------------------------
%  ENTAILMENT:      752        0    661 (113)      1
% ------------------------------------------------------
%  CONTRADICTION:     1      367     351 (56)      1
% ------------------------------------------------------
%  NEUTRAL:          24        4   2758 (217)      7
% ------------------------------------------------------
% Total #problems:  4927
% Accuracy (pure):  0.78831    (0.78833)
% Precision:        0.97474
% Recall (pure):    0.52437    (0.52486)
% ------------ STATS -------------
% Total #clTabperPrb:            1148
% Total #ruleApps for clTab:     13563
% Average #ruleApps for clTab:   11.81446
% true.
\end{document}